\documentclass[sigconf]{acmart}

\usepackage{epsfig}
\usepackage{graphicx}
\usepackage{amsmath}

\usepackage{array}
\usepackage{booktabs}
\usepackage{caption}
\usepackage{color}
\usepackage{multirow}
\usepackage{subfig}
\usepackage{ulem}
\usepackage{makecell}

\usepackage{amssymb}
\renewcommand\footnotetextcopyrightpermission[1]{}
\usepackage{fancyhdr}
\begin{document}

%%
%% The "title" command has an optional parameter,
%% allowing the author to define a "short title" to be used in page headers.
\title{Self-Paced Uncertainty Estimation for One-shot Person Re-Identification}

%%
%% The "author" command and its associated commands are used to define
%% the authors and their affiliations.
%% Of note is the shared affiliation of the first two authors, and the
%% "authornote" and "authornotemark" commands
%% used to denote shared contribution to the research.

%%
%% By default, the full list of authors will be used in the page
%% headers. Often, this list is too long, and will overlap
%% other information printed in the page headers. This command allows
%% the author to define a more concise list
%% of authors' names for this purpose.

\author{Yulin~Zhang, Bo~Ma, Longyao~Liu and Xin~Yi}
\affiliation{\institution{Beijing Laboratory of Intelligent Information Technology, School of Computer Science and Technology, Beijing Institute of Technology, Beijing 100081, China.}}
\email{{zhangyulin,bma000,roel_liu,yixin}@bit.edu.cn}
%%
 
%% article.
\begin{abstract}
	
  The one-shot Person Re-ID scenario faces two kinds of uncertainties when constructing the prediction model from $X$ to $Y$. The first is model uncertainty, which captures the noise of the parameters in DNNs due to a lack of training data. The second is data uncertainty, which can be divided into two sub-types: one is image noise, where severe occlusion and the complex background contain irrelevant information about the identity; the other is label noise, where mislabeled affects visual appearance learning. In this paper, to tackle these issues, we propose a novel Self-Paced Uncertainty Estimation Network (SPUE-Net) for one-shot Person Re-ID. By introducing a self-paced sampling strategy, our method can estimate the pseudo-labels of unlabeled samples iteratively to expand the labeled samples gradually and remove model uncertainty without extra supervision. We divide the  pseudo-label samples into two subsets to make the use of training samples more reasonable and effective. In addition, we apply a Co-operative learning method of local uncertainty estimation combined with determinacy estimation to achieve better hidden space feature mining and to improve the precision of selected pseudo-labeled samples, which reduces data uncertainty. Extensive comparative evaluation experiments on video-based and image-based datasets show that SPUE-Net has significant advantages over the state-of-the-art methods.
  
%  To promote better convergence, we decouple fused information. 
  
%  We decouple fused information  First, we propose a multi-to-one dehazing network to eliminate the haze distribution shift of images within the synthetic domain. Then, we conduct an inter-domain adaptation between the synthetic domain and the real domain based on the aligned synthetic features. 

%  the distribution shift within the synthetic domain  there are certain shortcomings. A distribution from the source domain can be converted to any distribution in the target domain without a constraint. We have proposed TSDN in this article
  
%  Recently, image dehazing task has achieved remarkable progress by convolutional neural network. However, intra-domain gap that haze distribution shift still exists in previous methods.  
%  intra-domain gap denotes  and inter-domain gap denotes domain shift.
  
%  However, those approaches train their model by coupled information, causing instability of the optimization. Besides, intra-domain gap that haze distribution shift within the domain 

%  mostly treat haze removal as a one-to-one problem and ignore the intra-domain gap. Therefore, haze distribution shift of the same scene images is not handled well. Also, dehazing models trained on the labeled synthetic datasets mostly suffer from performance degradation when tested on the unlabeled real datasets due to the inter-domain gap. Although some previous works apply translation network to bridge the synthetic domain and the real domain, the intra-domain gap still exists and affects the inter-domain adaptation. 

\end{abstract}

%%
%% Keywords. The author(s) should pick words that accurately describe
%% the work being presented. Separate the keywords with commas.

%% A "teaser" image appears between the author and affiliation
%% information and the body of the document, and typically spans the
%% page.
%\begin{teaserfigure}
%  \includegraphics[width=\textwidth]{sampleteaser}
%  \caption{Seattle Mariners at Spring Training, 2010.}
%  \Description{Enjoying the baseball game from the third-base
%  seats. Ichiro Suzuki preparing to bat.}
%  \label{fig:teaser}
%\end{teaserfigure}

%%
%% This command processes the author and affiliation and title
%% information and builds the first part of the formatted document.

\maketitle
\section{INTRODUCTION}\label{sec:introduction}

\begin{figure}[t] 
\begin{center}
%\fbox{\rule{0pt}{2in} \rule{1\linewidth}{0pt}}
\includegraphics[width=0.9\linewidth]{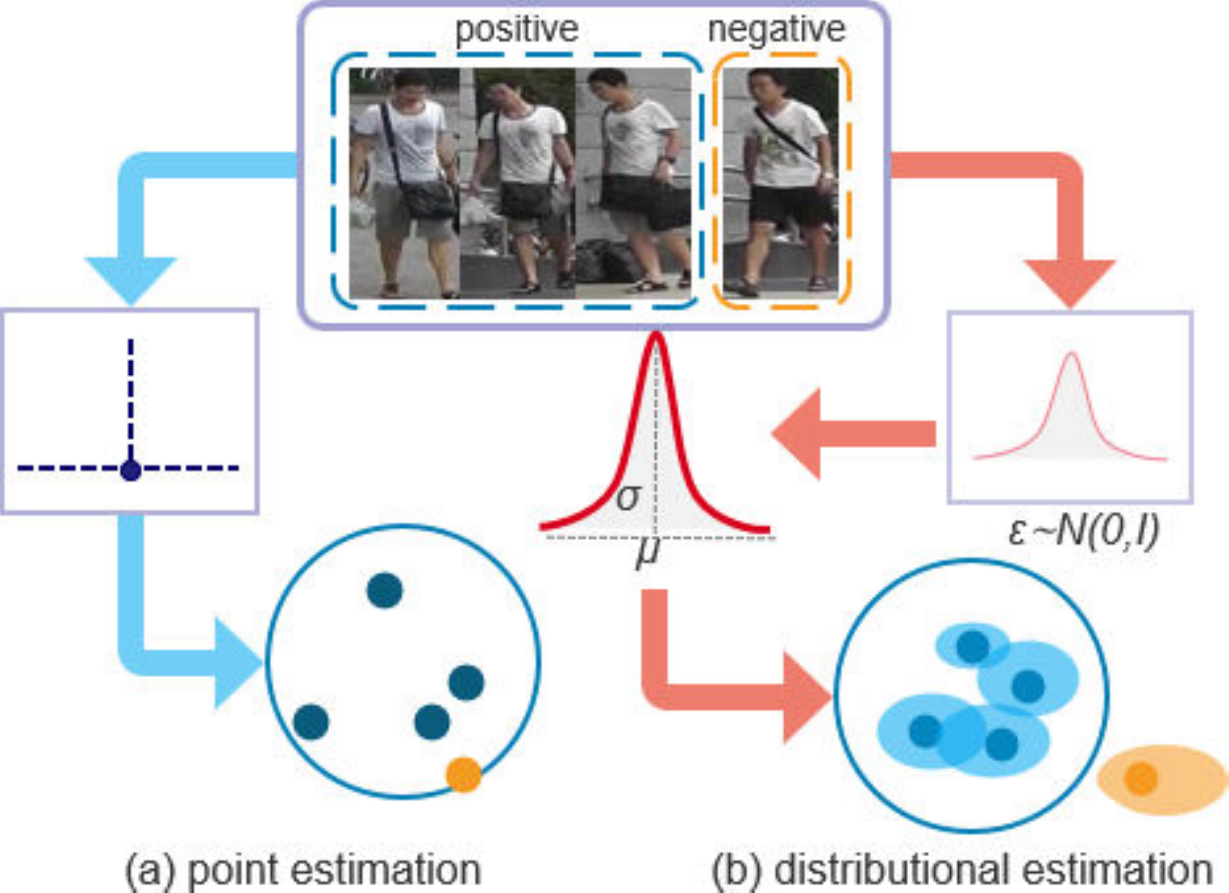}
\end{center}
\setlength{\abovecaptionskip}{1pt} 
   \caption{An illustration of the impact of the uncertainty estimation. The blue circle indicates a class label. Solid circles indicate the feature of samples, and solid ellipses indicate the uncertainty of samples.}
\label{fig:1}
\vspace{-0.2cm}

\end{figure}

One-shot Person Re-Identification(Re-ID) aims to reduce the cost of manual data labeling and the dependency of deep learning~\cite{he2016deep} on labeled data, that is, only one labeled sample for each identity is used for modeling. Due to the extreme lack of training data in the one-shot, the uncertainty of the Re-ID model parameters is called model uncertainty~\cite{kendall2017uncertainties,gal2016uncertainty,bender2018understanding}. This uncertainty can be eliminated if the data fed to the network is abundant. Therefore, in this work, we initialize a model using only one labeled sample for each identity and exploit distance-based metrics in the feature space to select pseudo-labeled samples as labeled samples from unlabeled samples by self-paced, which increases the number of labeled samples and removes the model uncertainty.

However, the image inherent noise can cause serious loss of discrimination information, which significantly hinders identity representation learning and reduces the precision of pseudo-label estimation. In particular, most Re-ID methods~\cite{zheng2019joint,wu2019distilled,wu2018exploit,li2019top,zhang2021one} represent each pedestrian image as a deterministic point embedded in the latent space. For the deterministic representation of the sample, the sample noise is not estimated, leading to misrecognition and mislabeled pseudo-labels. This is illustrated in Fig~\ref{fig:1}(a).

DistributionNet~\cite{yu2019robust} is the first work to consider noise estimation in person Re-ID. It describes each sample as a Gaussian distribution, rather than a deterministic point in the latent space. The mean of the distribution estimates the most likely feature values while the variance shows the uncertainty in the feature values. The work states explicitly how uncertainty affects feature learning, that is, samples with larger variances have less impact on the learned feature embedding space. The distributional estimation enhances the features and makes the same class of samples more compact, instead of overfitting samples of different identities, achieving better class separability, as illustrated in Fig~\ref{fig:1}(b). While being effective, DistributionNet learns the feature and noise in a supervised scenario. However, in the self-paced one-shot scenario, when mislabeled pseudo-label samples dominate the training, the model learns that the feature and variance can have larger shifts than the deterministic estimation, which seriously affects the performance of Re-ID.

For the first time, this work applies a novel Co-operative learning method of local uncertainty estimation combined with determinacy estimation to the self-paced one-shot person Re-ID, which is called Self-Paced Uncertainty Estimation Network (SPUE-Net). This effectively improves the feature affected by various sample noises without increasing the cost of labeling.

Specifically, we split the training samples into four parts, initial labeled samples, high-confidence pseudo-label samples (pseudo-label precision is relatively high), low-confidence pseudo-label samples (pseudo-label precision is relatively low), unlabeled samples. Uncertainty estimation is performed on the initial labeled samples and high-confidence pseudo-labeled samples. Determinacy estimation is performed on the low-confidence pseudo-label samples and unlabeled samples. We utilize universal index labels~\cite{wu2019progressive} for unlabeled samples. We discuss how the uncertainty estimation combined with determinacy estimation affects the model training, from the perspective of image noise. We provide insightful thinking that uncertainty estimation can improve the effect of classification model performance by relatively better quality samples.

Experimental results show that SPUE-Net achieves competitive performance on video-based MARS and DukeMTMC-VideoReID, as well as image-based datasets DukeMTMC-reID and Market-1501. 

The main contributions are summarized as follows:

\begin{itemize}
\item We propose a Self-Paced Uncertainty Estimation Network (SPUE-Net) to extract inherent identity representations for one-shot Person RE-ID. With the self-paced learning, the Co-operative learning method and subsets division strategy, our network is able to capture and remove the model uncertainty and data uncertainty without extra supervision information.

\item We apply a new Co-operative learning method of local uncertainty estimation combined with determinacy estimation to effectively improve the performance of the Re-ID affected by various sample noises without increasing the cost of labeling.

\item An effective subsets division strategy is proposed. Different subsets participate in different training methods, which reduces the influence of mislabeled samples on uncertainty estimation and realizes better-hidden space feature mining.

\item Experimental results on video-based datasets and image-based datasets show that SPUE-Net has significant advantages over the state-of-the-art methods.
\end{itemize}

%------------------------------------------------------------------------ 
\section{RELATED WORK}

\subsection{Uncertainty in deep learning}

Uncertainty learning can improve the performance of deep learning models, which has attracted significant attention from researchers~\cite{kendall2017uncertainties,gal2016uncertainty,kendall2018multi,depeweg2018decomposition}. There are two main types of uncertainty, ie, Epistemic uncertainty and Aleatoric uncertainty. Epistemic uncertainty~\cite{kendall2017uncertainties} is uncertainty in the model, capturing the noise of parameters  in deep neural networks due to a lack of training data~\cite{kendall2017uncertainties}. Aleatoric uncertainty~\cite{kendall2017uncertainties,chang2020data} captures uncertainty about the information that the data cannot explain, divided into two sub-categories, Heteroscedastic uncertainty and Homoscedastic uncertainty. Heteroscedastic uncertainty depends on the input data~\cite{kendall2017uncertainties}. While homoscedastic uncertainty is Task-dependent, which is a quantity that stays constant for all input data but varies between different tasks~\cite{kendall2017uncertainties,kendall2018multi}. 

Uncertainty has not been applied by vision application until recent years. The uncertainty method is first applied to modern DNN-based vision tasks in~\cite{kendall2017uncertainties}. This method models regression tasks and per-pixel semantic segmentation. In this paper, a novel Bayesian deep learning framework is presented to learn a mapping to data uncertainty from the input samples, which is composed on top of model uncertainty. This network improves the baseline. Besides, object detection~\cite{he2019bounding,choi2019gaussian}, semantic segmentation~\cite{kendall2017uncertainties,kendall2018multi,zheng2020rectifying}, and face recognition~\cite{chang2020data,zafar2019face,shi2019probabilistic} tasks introduce uncertainty learning to enhance the performance of the model. In classification tasks, some studies~\cite{chang2020data,yu2019robust} use uncertainty learning for analysis and model each ID image in latent space as a Gaussian distribution.

\subsection{Person Re-Identification}
Person Re-ID is a cross-view Person retrieval task, which needs to accurately label each pedestrian from multiple camera views. The state-of-the-art person RE-ID algorithms based on deep learning techniques are developed under supervised learning~\cite{zheng2015scalable,lin2019improving,kalayeh2018human,ristani2018features, zhang2019densely,zheng2018pedestrian,wu2018and}. However, the availability of labeled training samples is limited by the sharply increased cost of manual efforts. This dilemma has prompted people to study one-shot person Re-ID~\cite{wu2018exploit,yang2017enhancing,bak2017one,wu2019progressive,zhang2021one} and establish effective algorithms to reduce the cost of labeling. The one-shot is that there is only one labeled sample for each identity. These methods can make better use of unlabeled data without manual intervention.

Most Re-ID methods~\cite{zheng2019joint,xu2018attention,wu2019distilled,zheng2015scalable,lin2019improving,kalayeh2018human,ristani2018features, zhang2019densely,ye2017dynamic,zhong2017re} represent each pedestrian image as a deterministic point embedding in the latent space. The pedestrian features learned by the embedding model could be ambiguous or may not even be present in the input pedestrian, leading to noisy representations. The ideal embedding features can best represent pedestrian identity information and will not be interfered with by information unrelated to pedestrian identity. However, the commonly used datasets for Person Re-ID~\cite{zheng2015scalable,lin2019improving,zheng2016mars,wu2018exploit} are acquired in a real scenario. These pedestrian images inevitably carry information that has nothing to do with the identity of pedestrians. As the input data contains lots of noise, Data-dependent uncertainty modeling is essential.

%Using Photorealistic Face Synthesis and Domain Adaptation to
%Improve Facial Expression Analysis
\subsection{Self-Paced learning}

%During the past few decades, significant progress has been made on FER. A variety of FER methods have been proposed, where these methods can be roughly classified into two categories: hand-crafted features based methods and CNN features based methods.

Curriculum learning~\cite{bengio2009curriculum} and Self-paced learning~\cite{kumar2010self-paced,lee2011learning,zhang2017co-saliency,jiang2014self-paced} are originally proposed for solving non-convex optimization problems. Bengio et al.~\cite{bengio2009curriculum} suggest that curriculum learning can be seen as a continuation method. Self-paced learning is developed based on curriculum learning. Kollerteam et al.~\cite{kumar2010self-paced} modeled this idea into a mathematical expression with the theoretical foundation. The core idea of curriculum learning and self-paced learning is to simulate the cognitive mechanism of humans. Both modes process samples in a meaningful order. First, they learn a simple knowledge structure; secondly, the difficulty of learning is continuously increased; finally, more complex and professional knowledge is acquired. We introduce the core idea of self-paced learning into the one-shot Person Re-ID scenario to discover more data information and to train a robust feature classifier for the improvement of the model performance. Our approach also has a similar flavor to active learning~\cite{settles2009active,johnson2008active} and co-training~\cite{blum1998combining,han2020self}. We select the most confidently pseudo-labeled samples by alternately training classifiers to learn for the next training.
%------------------------------------------------------------------------

\section{SPUE-Net}
\begin{figure*}[t] 

\begin{center}
%\fbox{\rule{0pt}{2in} \rule{1\linewidth}{0pt}}
\includegraphics[width=1\linewidth]{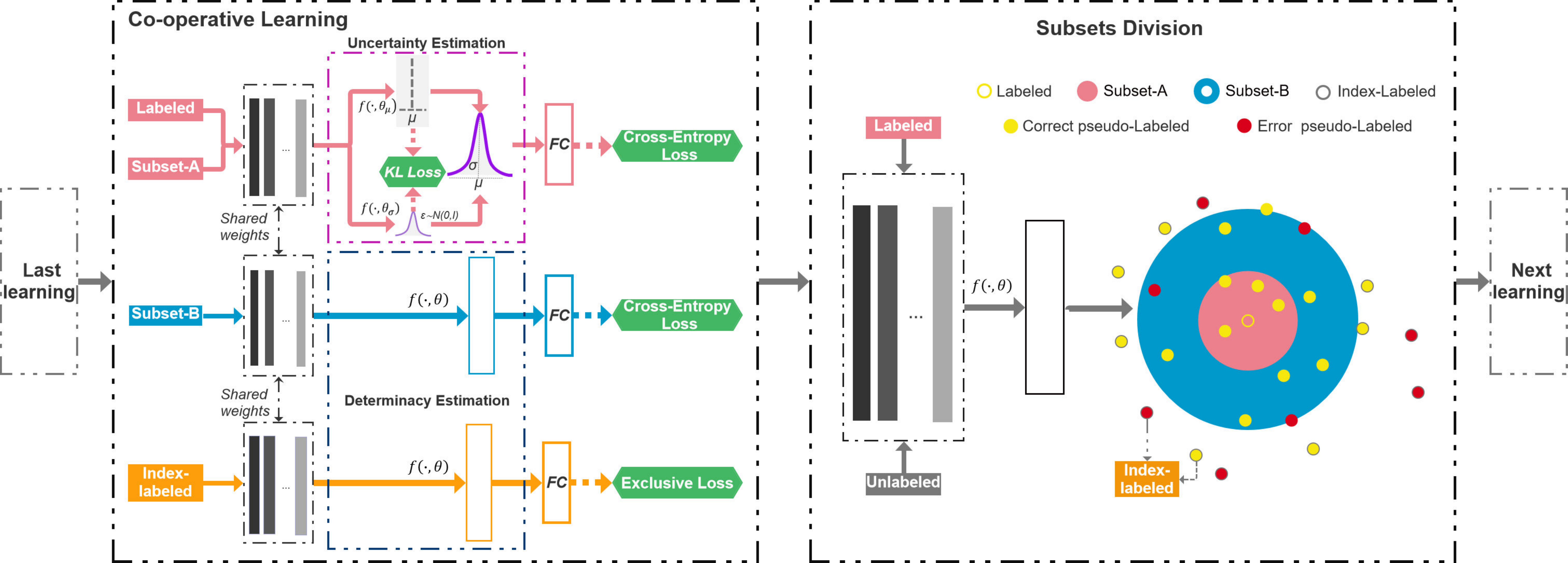}
\end{center}
\setlength{\abovecaptionskip}{1pt} 
   \caption{A schematic Overview of SPUE-Net.}
\label{fig:2}
\vspace{-0.2cm}
\end{figure*}
%-------------------------------------------------------------------------

\subsection{Overview}\label{3.1}

The overview of the proposed method is given in Figure.\ref{fig:2}. The specific process is as follows: First, we optimize the network by the Co-operative learning method of local uncertainty estimation combined with determinacy estimation on the four kinds of sample splits (the labeled samples, pseudo-labeled Subset-$\mathcal{A}$ samples, pseudo-labeled Subset-$\mathcal{B}$ samples, and index-labeled samples); Second, a few reliable pseudo-labeled samples are selected according to the estimation of pseudo-labels for all unlabeled samples in the feature space. Next, we divide the selected pseudo-labeled samples into Subset-$\mathcal{A}$ and Subset-$\mathcal{B}$ by subsets division strategy based on distance confidence. Specifically, Subset-$\mathcal{A}$ and Subset-$\mathcal{B}$ are empty sets in the first iteration. We continually expand the number of selected pseudo-label samples and unselected unlabeled samples are assigned index labels as index data to joint train the model. We utilize the initial labeled samples and the Subset-$\mathcal{A}$ samples to learn the features and variance of the sample simultaneously by Cross-entropy classification loss and KL loss. We use Cross-entropy classification loss and exclusive loss to constrain Subsets-$\mathcal{B}$ and index data, respectively.

\subsection{Problem formulation}\label{3.4}

For one-shot setting, there is a labeled dataset labeled set $L=$ $\left\{\left(x_{1}^{L}, y_{1}^{L}\right),\left(x_{2}^{L}, y_{2}^{L}\right), \cdots\left(x_{n}^{L}, y_{n}^{L}\right)\right\}$ with $n$ identities and an unlabeled dataset $U=\left\{x_{i}^{U}\right\}_{i=1}^{m}$ including $m$ images. These samples are used to train a feature extractor $f\left(x_{i}, \theta\right)$ which parameterized by $\theta$ and an identity classifier $\phi(\cdot, \delta)$ which parameterized by $\delta$ in the way of identity classification. In the test stage, after the deep model $f(\cdot, \theta)$ is obtained, the similarity is computed according to the Euclidean Distance in the feature space $\left\|f\left(x^{q}, \theta\right)-f\left(x^{g}, \theta\right)\right\|$ between the query samples $x^{q}$ and gallery samples $x^{g}$.  The result of the query is a ranked list of all gallery data. We estimate the identity label $\hat{y}_{i}$ of the unlabeled samples $x_{i}^{U}$ to expand the labeled sample and select reliable pseudo-label samples for the Re-ID learning. We define $\mathcal{A}^{t}$, $\mathcal{B}^{t}$, $P^{t}$ and $I^{t}$ as the Subset-$\mathcal{A}$ samples, Subset-$\mathcal{B}$ samples, selected pseudo-label samples and the index-labeled of $t$-th step iteration, respectively.

%-------------------------------------------------------------------------

\subsection{The Co-operative learning method}\label{3.2} 

We first introduce the iterative steps of SPUE-Net. We utilize samples (initial labeled samples $L$, $\mathcal{A}^{t}$, $\mathcal{B}^{t}$) with identity labels, and $I^{t}$ with index labels for training at the $t$-th iteration. We utilize $L$ and $\mathcal{A}^{t}$ to learn an identity classifier with their identity label by uncertainty estimation. For the $\mathcal{B}^{t}$, their pseudo-labels are relatively unreliable compared to $\mathcal{A}^{t}$ and may harm the training of uncertainty estimation in the network. Therefore, we apply determinacy estimation on $\mathcal{B}^{t}$ and use the exclusive loss \cite{wu2019progressive} on $I^{t}$ to co-operative optimize the network.

\textbf{Determinacy estimation} The method of deterministic feature embedding on $\mathcal{B}^{t}$ and index-labeled samples $I^{t}$ is a continuous mapping space from $X$ to $Y$. Here $X$ is a continuous image space, and $Y$ is a discrete identifier. Each sample $x_{i}^{B} \in B^{t}$ or $x_{i}^{I} \in I^{t}$is represented as an embedding $z_{i}$ in the latent space. For the$\mathcal{B}^{t}$ samples, we apply an identity classifer $\phi(\cdot, \delta)$ on their deterministic feature embedding $f(\cdot, \theta)$ and optimize the network by comparing the expected probability (predicted identity) with real probability (pseudo-label $\hat{y}_{i}$). We have the following objective function.

\begin{equation}
\mathcal{L}_{D E}=-\frac{1}{N} \sum_{i=1}^{N} q\left(\hat{y}_{i}\right) \log \frac{\exp \left(\phi\left(f\left(x_{i}^{B}, \theta\right), \delta\right)\right)}{\sum_{n=1}^{N} \exp \left(\phi\left(f\left(x_{i}^{B}, \theta\right), \delta\right)\right)}
\label{e5}
\vspace{-0.2cm}
\end{equation}
where $N$ is the class number, $q\left(\hat{y}_{i}\right)$ represents the indicator variable.  Minimizing this cross-entropy loss $ L_{D E}$ makes the model output closer to the predicted result.

For index-labeled samples $I^{t}$, we utilize the Exclusive loss \cite{wu2019progressive} to learn a discriminative embedding and to extend the distance between each samples $I^{t}$ in the feature space, as Eq.(\ref{e2}).

\vspace{-0.2cm}
\begin{equation}
\mathcal{L}_{E X}=-\log \left(\frac{\exp \left\|W_{y_{i}}\right\|\left\|x_{i}^{U}\right\| \cos \left(\theta_{y_{i}}\right)}{\sum_{j=1}^{M} \exp \left\|W_{j}\right\|\left\|x_{j}^{U}\right\| \cos \left(\theta_{j}\right)}\right)
 \label{e2}
\end{equation}
\vspace{-0.2cm}
where $x_{i}^{U} \neq x_{j}^{U}$ and $x_{i}^{U}, x_{j}^{U} \in I^{t}$.

\textbf{Uncettainty estimation} If $x_{i} \in L$ or $x_{i} \in \mathcal{A}^{t}$ has an ideal embedding $f\left(x_{i}, \theta\right)$, which only has identity correlation information. Typically it is inaccurate to represent an identity embedding $f\left(x_{i}, \theta\right)$ by only a single pedestrian sample $x_{i}$ and more samples need to be collected. The pedestrian samples are acquired in a real scenario, which inevitably carries irrelevant identity information. The extraction of pedestrian ideal features is affected by irrelevant identity information. Therefore, this feature mapping space contains image noise. If we hypothesize that the noise in the image is additive and obeys a Gaussian distribution, whose mean is zero, and the variance depends on the sample $x_{i}$. Therefore, the sample $x_{i}$ can be represented as $z_{i}=f\left(x_{i}\right)+n\left(x_{i}\right)$ in the latent space during deep learning, where $n\left(x_{i}\right)$ is the uncertainty information of the sample $x_{i}$.

We estimate a distribution $p\left(z_{i} \mid x_{i}\right)$ in the latent space to represent the potential appearance of sample $x_{i}$. The representation $z_{i}$ in latent space is defined as a Gaussian distribution.
\vspace{-0.2cm}
\begin{equation}
p\left(z_{i} \mid x_{i}\right)=\mathcal{N}\left(z_{i}, \mu_{i}, \sigma_{i}^{2} I\right)
 \label{e3}
\end{equation}
\vspace{-0.2cm}

We establish two independent branches in the penultimate feature extraction layer of the network for generating $\mu_{i}$ and $\sigma_{i}$: $\mu_{i}=f\left(x_{i}, \theta_{\mu}\right)$ and $\left.\sigma_{i}=f\left(x_{i}, \theta_{\sigma}\right)\right\}$. To reduce the pedestrian representation complexity, we only use the diagonal covariance matrix. Mean $\mu_{i}$ can be regarded as the most likely pedestrian feature of the input sample, and uncertainty $\sigma_{i}$ can be regarded as the confidence of the model along each feature dimension. The representation of each sample $z_{i}$ is a random embedding sampled from $\mathcal{N}\left(z_{i}, \mu_{i}, \sigma_{i}^{2} I\right)$ in the latent space. When the random sample $z_{i}$ is fed to the network, the sampling operation is not differentiable, which prevents the error back-propagation to the preceding layers during the model training. To address this problem, we use a reparameterization trick ~\cite{kingma2013auto} during sampling to let those layers benefit from the random samples $z_{i}$. We first draw a sample $\varepsilon$ from the normal distribution $\mathcal{N}(0, I)$ with zero mean and unit covariance, instead of directly sampling from $\mathcal{N}\left(\mu_{i}, \sigma_{i}^{2}\right)$. Then we get the sample $\mathcal{R}_{i}$ by computing Eq.(\ref{e4}), which is equivalent to the sample representation.
\vspace{-0.2cm}
\begin{equation} 
\mathcal{R}_{i}=\mu_{i}+\varepsilon \sigma_{i}, \quad \varepsilon \sim \mathcal{N}(0, I)
 \label{e4}
\end{equation}
\vspace{-0.2cm}

In this way, this sampling operation does not need gradient descent and the entire model is trainable. We input the final representation $\mathcal{R}_{i}$ to the identity classifier $\phi(\cdot, \delta)$ and utilize the cross-entropy loss for classification training, as formulated in Eq.(\ref{e6})

\begin{equation}
\mathcal{L}_{C E}=-\frac{1}{N} \sum_{i=1}^{N} q\left(y_{i}\right) \log \frac{\exp \left(\phi\left(\mathcal{R}_{i}, \delta\right)\right)}{\sum_{n=1}^{N} \exp \left(\phi\left(\mathcal{R}_{i}, \delta\right)\right)}
 \label{e6} 
\end{equation}
where $q\left(y_{i}\right)$ represents the indicator variable.

During the training stage, since each identity embedding feature $\mu_{i}$ is corroded by uncertainty $\sigma_{i}$, the trivial solution of variance gradually drops to zero. To prevent this problem, we suppress the unstable factors in $\mathcal{R}_{i}$ so that the Eq.(\ref{e6}) can converge eventually. In this case, the $\mathcal{R}_{i}$ is re-expressed as $\mathcal{R}_{i}=\mu_{i}+c_{ }$, which degrades into an original feature representation. Influenced by deep variational information bottleneck (VIB) theory~\cite{alemi2018uncertainty,alemi2016deep,Zhang_2019_ICCV}, the distribution $\mathcal{N}\left(\mu_{i}, \sigma_{i}^{2}\right)$ approaches the normal distribution $\mathcal{N}(0, I)$ during the optimization by adding a regularization term to encourage the model to learn the feature uncertainty. These two distributions are measured by Kullback-Leibler divergence (KLD), as formulated in Eq.(\ref{e7}).

\begin{equation}
\begin{aligned}
\mathcal{L}_{K L} &=K L\left[N\left(z_{i} \mid \mu_{i}, \sigma_{i}^{2}\right) \| N(\varepsilon \mid 0, I)\right] \\
&=\frac{1}{2} \sum_{i=1}^{d}\left(\mu_{i}+\sigma_{i}^{2}-\log \sigma_{i}^{2}-1\right)
\end{aligned}
 \label{e7}
\end{equation}
where $d$ is the dimension of the hidden variable $z_{i}$. The loss $\mathcal{L}_{KL}$ balance the classification loss of Eq.(\ref{e6}). To acquire the outstanding features embedding by the uncertainty estimation branch, we employ the joint training of softmax loss and KL loss to optimize the model. Therefore, the cost function of uncertainty estimation Eq.(\ref{e8}) is as follows:

\begin{equation}
\mathcal{L}_{U E}=\mathcal{L}_{C E}+\lambda \mathcal{L}_{K L}
\label{e8}
\end{equation}
where the hyper-parameter $\lambda$ is used for balancing the Cross-Entropy loss and KL loss functions.
%-------------------------------------------------------------------------

\textbf{The SPUE-Net objective function} Considering the four data splits, in the $(t+1)$-$\mathrm{th}$ iteration of the model, we formulate the final objective function as Eq.(\ref{e9}):
\begin{equation}
\begin{array}{c}
\mathcal{L}_{S P U E} =\gamma \mathcal{L}_{U E}^{L}+ \\
 \mathrm{s}^{t}\gamma\left[\alpha^{t} \mathcal{L}_{U E}^{\mathcal{A}}+\left(1-\alpha^{t}\right) \mathcal{L}_{D E} \right]+ \\
\left(1-\mathrm{s}^{t}\right)(1-\gamma) \mathcal{L}_{E X}
\end{array}
 \label{e9}
\end{equation}
where $\alpha^{t} \in\{0,1\}$ is the selection indicator for the Subset-$\mathcal{A}$ and $s^{t} \in\{0,1\}$ is the selection indicator for the unlabeled sample. The first part is the uncertainty estimation loss $\mathcal{L}_{UE}^{L}$ on the labeled sample $L$. The second part includes the uncertainty loss $\mathcal{L}_{UE}^{\mathcal{A}}$ on the divided Subset-$\mathcal{A}$ samples ${\mathcal{A}^{t+1}}$. The third one is the determinacy estimation loss $\mathcal{L}_{DE}$ on the divided Subset-$\mathcal{B}$ subset ${\mathcal{B}^{t+1}}$ . The last one is the exclusive loss $\mathcal{L}_{EX}$ on index-labeled data $I^{t+1}$. The hyper-parameter $\gamma$ is set to adjust the proportion of the loss function between the index label sample and the identity label sample. The model does not complete the training until all unlabeled samples are selected into the pseudo-labeled dataset.
%-------------------------------------------------------------------------

\subsection{The effective subsets division}\label{3.3}
\textbf{Selected pseudo-label samples} Subset division is performed on the selected pseudo-label samples, and the precision of the pseudo-label samples is crucial to the subset division. Previous works feed unlabeled samples into the classification model to estimate pseudo labels. The overfitting phenomenon of the classification model cannot accurately predict the class of each unlabeled sample. To address this, we employ the Nearest Neighbors (NN) classifier to estimate the label of unlabeled data. The Euclidean distance is utilized between unlabeled samples $x_i^U$ and initial labeled samples $x_i^L $ in the feature space as the confidence of label estimation.
\begin{equation}
\begin{array}{l}
d\left(x_{i}^{U}, \theta\right)=\arg \min \left\|f\left(x_{i}^{U}, \theta\right)-f\left(x_{i}^{L}, \theta\right)\right\| \\         
\end{array}
\label{e11}
\end{equation}
where $d\left(x_{i}^{U}, \theta\right)$ denotes the result of the label estimation. We select partial unlabeled samples of top confidence as pseudo-label samples $P^t$ and set a step size called the Expansion Rate (ER) to determine the number of pseudo labels selected in each iteration. In the $t$-th step iteration, the highest reliable pseudo-labeled samples of  $ER \times t$  are selected according to Eq.(\ref{e12}).

\begin{equation}
\begin{array}{l}
 
P^{t}=\arg \min _{E R \times t} \sum_{i=n+1}^{n+m} p_{i} d\left(x_{i}^{U}, \theta\right)
\end{array}
\label{e12}
\end{equation}
where $P^t$ is the vertical concatenation of all $p_i$.

\textbf{Subsets division} Due to the lack of initial labeled samples $L$, the model has an over-fitting phenomenon. That is, the selected pseudo-label samples $P^t$ by Eq.(\ref{e12}) and Eq.(\ref{e12}), there are still a large number of mislabeled samples, and the precision of the pseudo-label samples gradually decreases as the confidence decreases. This study is also described in ~\cite{zhang2021one}.

To reduce the impact of pseudo-labeled noise on data uncertainty estimation, we divide the selected pseudo-label samples $P^t$ into two subsets: Subset-$\mathcal{A}$ and Subset-$\mathcal{B}$.

\begin{equation}
\begin{array}{c}

\mathcal{A}^{t}= \arg \min _{\alpha \times E R \times t} \sum_{i=n+1}^{n+m} p_{i} d\left(x_{i}^{U}, \theta\right) \\

\end{array}
\label{e13}
\end{equation}

\begin{equation}
\begin{array}{c}

\mathcal{B}^{t}=\arg \max _{\left(1-\alpha\right)\times E R \times t}\left[\arg \min _{E R \times t} \sum_{i=n+1}^{n+m} p_{i} d\left(x_{i}^{U}, \theta\right)\right]
\end{array}
\label{e14}
\end{equation}
where $\alpha$ is a hyperparameter to adjust the proportion between Subset-$\mathcal{A}$ and Subset-$\mathcal{B}$. The selection of hyperparameter $\alpha$ values depends on the sample quality. The smaller the hyperparameter $\alpha$, the smaller the weight of the uncertainty estimation. However, the quality of samples in Subset-$\mathcal{A}$ also decreases simultaneously, and the negative impact of uncertainty gradually increases. Through multiple experiments, we selected 0.3 for the hyperparameter, which has a relatively good influence on the model. See the experiment~\ref{sec:fes} for details.

%-------------------------------------------------------------------------

%------------------------------------------------------------------------

%------------------------------------------------------------------------
\section{Experiments}

\begin{table*}[]
  \centering
  \Large
\vspace{-0.2cm}
\setlength{\tabcolsep}{0.8mm}
\caption{Comparison with the state-of-the-art methods on Market-1501 and DukeMTMC-reID. All comparison methods are based on the same backbone model ResNet50. The baseline(initial) is the initial model trained on our proposed method. ER is the expansion rate that the expansion speed of sample label estimation. The best results are in black boldface font.}

\begin{tabular}{p{5pt}c!{\vrule width1.3pt}ccccc!{\vrule width1.3pt}ccccc}
\Xhline{1.3pt}
\multicolumn{2}{c!{\vrule width1.3pt}}{\multirow{2}{*}{methods}}                               & \multicolumn{5}{c!{\vrule width1.3pt}}{Market-1501}                                              & \multicolumn{5}{c}{DukeMTMC-reID}                                            \\ \cline{3-12} 
\multicolumn{2}{c!{\vrule width1.3pt}}{}                          & mAP           & Rank-1        & Rank-5        & Rank-10       & Rank-20       & mAP           & Rank-1        & Rank-5        & Rank-10       & Rank-20       \\ \Xhline{1.1pt}
\multicolumn{2}{c!{\vrule width1.3pt}}{Baseline(initial)}                                      & 10.2          & 28.3          & 45.5          & 52.3          & 60.1          & 8.9           & 21.4          & 33.4          & 39.5          & 45.1          \\ \Xhline{1.2pt}
\multicolumn{1}{c|}{\multirow{2}{*}{ER=30\%}} & PL\cite{wu2019progressive} & 13.4          & 35.5          & 52.8          & 60.5          & 68.6          & 11.1          & 23.3          & 35.7          & 42.2          & 48.0          \\
\multicolumn{1}{c|}{}                         & Ours                       & \textbf{16.2} & \textbf{38.5} & \textbf{56.4} & \textbf{63.8} & \textbf{72.7} & \textbf{13.8} & \textbf{28.0} & \textbf{42.2} & \textbf{48.4} & \textbf{54.6} \\ \hline
\multicolumn{1}{c|}{\multirow{2}{*}{ER=20\%}} & PL\cite{wu2019progressive} & 17.4          & 41.4          & 59.6          & 66.4          & 73.5          & 15.1          & 30.0          & 43.4          & 49.2          & 54.8          \\
\multicolumn{1}{c|}{}                         & Ours                       & \textbf{19.6} & \textbf{45.0} & \textbf{61.1} & \textbf{68.4} & \textbf{74.9} & \textbf{20.3} & \textbf{38.1} & \textbf{52.2} & \textbf{57.6} & \textbf{63.0} \\ \hline
\multicolumn{1}{c|}{\multirow{2}{*}{ER=15\%}} & PL\cite{wu2019progressive} & 19.2          & 44.8          & 61.8          & 69.1          & 76.1          & 18.2          & 35.1          & 49.1          & 54.3          & 60.0          \\
\multicolumn{1}{c|}{}                         & Ours                       & \textbf{23.6} & \textbf{51.0} & \textbf{66.8} & \textbf{73.2} & \textbf{79.2} & \textbf{22.3} & \textbf{40.6} & \textbf{54.6} & \textbf{60.2} & \textbf{65.4} \\ \hline
\multicolumn{1}{c|}{\multirow{2}{*}{ER=10\%}} & PL\cite{wu2019progressive} & 23.2          & 51.5          & 66.8          & 73.6          & 79.6          & 21.8          & 40.5          & 53.9          & 60.2          & 65.5          \\
\multicolumn{1}{c|}{}                         & Ours                       & \textbf{28.3} & \textbf{57.5} & \textbf{73.4} & \textbf{79.6} & \textbf{84.6} & \textbf{25.0} & \textbf{45.6} & \textbf{59.7} & \textbf{65.0} & \textbf{69.5} \\ \hline
\multicolumn{1}{c|}{\multirow{2}{*}{ER=5\%}}  & PL\cite{wu2019progressive} & 26.2          & 55.8          & 72.3          & 78.4          & 83.5          & \textbf{28.5} & \textbf{48.8} & \textbf{63.4} & \textbf{68.4} & \textbf{73.1} \\
\multicolumn{1}{c|}{}                         & Ours                       & \textbf{30.5} & \textbf{60.4} & \textbf{76.0} & \textbf{81.1} & \textbf{85.7} & 27.6          & 46.3          & 61.0          & 67.5          & 71.0        \\\Xhline{1.3pt}

\end{tabular}
\
\setlength{\abovecaptionskip}{1pt}

\label{tab Table 1}
\vspace{-0.2cm}
\end{table*}

%-------------------------------------------------------------------------
\subsection{Experimental Datasets}
\label{sec:database}
We conduct experiments to evaluate the proposed Self-Paced Uncertainty Network on image-based datasets (Market1501, DukeMTMC-reID) and video-based datasets (MARS, DukeMTMC-VideoReID).

\textbf{Market1501}~\cite{zheng2015scalable} is an image dataset covering 32,668 labeled images of 1,501 identities collected from 6 camera views. Thereinto 12,936 images from 751 identities are used for training and 19,732 images from 750 identities for testing.

%-------------------------------------------------------------------------

\textbf{DukeMTMC-reID} contains 36411 labeled images of 1,812 identities from 8 camera views. We split the dataset into two non-overlapping parts according to the setup in~\cite{ristani2016performance}: 16,522 images from 702 identities for training and 19989 images from other 702 identities for testing.

%-------------------------------------------------------------------------

\textbf{MARS}~\cite{wu2018exploit} has 17,503 tracklets captured by 6 cameras, containing 1,261 identities and 3,248  distractor tracklets, of which 625 identities for training and 636 identities for testing.

%-------------------------------------------------------------------------
\textbf{DukeMTMC-VideoReID}~\cite{zheng2016mars} is a video-based dataset. It includes 2,196 videos of 702 identities for training and 2,636 videos of 702 identities for testing. 408 identities are used as distractors. Each video contains images sampled every 12 frames.

\subsection{Experimental Setups}
\label{sec:implementation}

\subsubsection{Performance Metrics} We use two evaluation metrics: mean Average Precision (mAP) and the Cumulative Matching Characteristics (CMC) curve to evaluate the proposed effectiveness of model. As an object retrieval problem, the mAP represents the recall rate, and it is the mean value of the average accuracy of all queries. And the CMC score reflects retrieval accuracy. We report the Rank-1, Rank-5, Rank-10, Rank-20 scores to represent the CMC curve.

%-------------------------------------------------------------------------

\subsubsection{Dataset setting} We apply the~\cite{liu2017stepwise} protocol directly to one-shot Person RE-ID. In image-based and video-based datasets, we randomly select an image/tracklet under camera 1 for each identity pedestrian as the labeled data. If an identity has no sample under camera 1, then it is postponed to camera 2 to select the sample, and so on. Make sure that each identity has a labeled example.

%-------------------------------------------------------------------------

\subsubsection{Implementation Details} To implement our algorithm, we adopt ResNet-50 pre-trained on ImageNet~\cite{krizhevsky2012imagenet} with the last classification layer removed as our basic feature embedding model for feature extraction. To make better use of unlabeled data, we use the fully connected layer and a classified layer with batch normalization~\cite{ioffe2015batch} at the end of the feature embedding model. 
The hyper-parameter $\lambda$ is set to 0.01 to balance the Cross-Entropy loss and KL loss functions. The hyper-parameter $\gamma$ is set to 0.8 and the hyper-parameter $\alpha$ is set to $0.3$ to control the final loss of model training. For the optimization model, we employ Stochastic Gradient Descent (SGD) with a weight decay of $0.0005$ and a momentum of $0.5$ to optimize the loss function parameters. The batch size is set to $16$ at Marker1501 and DukeMTMC-reID image datasets and is set to $8$ at MARS and DukeMTMC-VideoReID video datasets. There are $70$ epochs in each iterative. The initial learning rate is set to $0.1$ and the learning rate is reduced to $0.01$ when epoch$>55$. Our model is trained end-to-end on a single Nvidia GeForce GTX 1080Ti with 12 GB memory, and the total time is about $10$ hours when the expansion rate is 10\% at the Marker1501 dataset.

\subsection{Comparison to the state-of-the-art}
\label{sec:cmp_state}

We compare our proposed SPUE-Net with the state-of-the-art, and the initial model uses the same one-shot sample of each identity. Besides, we compare our proposed method with DGM and Stepwiseon video-based datasets. The specific experiment design is consistent with PL\cite{wu2019progressive}. To make it fair, our labeled data Expansion Rates $ER$ is set to the same 5\%, 10\%, 15\%, 20\%, and 30\% as the PL. First of all, Table~\ref{tab Table 1} shows that the rank-1 accuracy of our proposed method has exceeded our initial model 32.1, 24.9, 24.9, and19.9 points on the Market-1501, DukeMTMC-reID, MARS, and DukeMTMC-VideoReID dataset, respectively. 

The SPUE-Net increases rank-1 accuracy by 3.0, 3.6, 6.2, 6.0, and 4.8 points at the different expansion rates on the Market-1501 dataset. Simultaneously, rank-1 accuracy is increased by 4.7, 8.1, 4.5, 5.1, and -2.5 points on the DukeMTMC-reID. We notice that the performance of the model has a certain decrease relative to the baseline PL\cite{wu2019progressive} when ER=5\% on the DukeMTMC-reID dataset. We provide insightful analysis of the causes of experimental performance degradation. On the video-based datasets, our proposed method increases rank-1 accuracy by 0.2, 1.5, 1.8, 3.0, and 3.9 points at the different expansion rates on the MARS dataset compared with PL\cite{wu2019progressive}. Our proposed method increases rank-1 accuracy by 5.0, 1.2, 2.8, 0.3, and 0.5 points at the different expansion rates on the DukeMTMC-VideoReID dataset compared with EUG\cite{wu2018exploit}. The performance of SPUE-Net is significantly improved in comparison with PL\cite{wu2019progressive} and EUG\cite{wu2018exploit}.

\begin{table*}[]
  \centering
  \Large

\setlength{\fboxsep}{0pt}
\setlength{\tabcolsep}{0.9mm}
\caption{Tables 1 Comparison with the state-of-the-art methods on MARS and DukeMTMC-VideoReID. All comparison methods are based on the same backbone model ResNet50. ER is the expansion rate that the expansion speed of sample label estimation. The best results are in black boldface font.}

\begin{tabular}{cc!{\vrule width1.3pt}ccccc!{\vrule width1.3pt}ccccc}
\Xhline{1.3pt}
\multicolumn{2}{c!{\vrule width1.3pt}}{\multirow{2}{*}{methods}}               & \multicolumn{5}{c!{\vrule width1.3pt}}{MARS}                                                     & \multicolumn{5}{c}{DukeMTMC-VideoReID}                                       \\ \cline{3-12} 
\multicolumn{2}{c!{\vrule width1.3pt}}{}         & mAP           & Rank-1        & Rank-5        & Rank-10       & Rank-20       & mAP           & Rank-1        & Rank-5        & Rank-10       & Rank-20       \\ \Xhline{1.2pt}
\multicolumn{2}{c!{\vrule width1.3pt}}{Baseline(initial)}                     & 16.8          & 37.6 & 52.5          & 57.6         & 63.8          & 45.8          & 53.4          & 70.4          & 74.8          & 80.3          \\
\multicolumn{2}{c!{\vrule width1.3pt}}{DGM+IDE\cite{ye2017dynamic}}                               & 16.9          & 36.8          & 54.0          & 59.6          & 68.5          & 33.6          & 42.4          & 57.9          & 63.8          & 69.3          \\
\multicolumn{2}{c!{\vrule width1.3pt}}{Stepwise\cite{liu2017stepwise}}                              & 19.7          & 41.2          & 55.6          & 62.2          & 66.8          & 46.8          & 56.3          & 70.4          & 74.6          & 79.2          \\ \Xhline{1.2pt}
\multicolumn{1}{c|}{} 
                             & EUG\cite{wu2018exploit}     & 21.1          & 42.8          & 56.5          & -             & 67.2          & 54.6          & 63.8          & 78.6          & -             & 87.0          \\
\multicolumn{1}{c|}{ER=30\%}                      & PL\cite{wu2019progressive}  & 22.1          & 44.5          & 58.7          & 65.7          & 70.6          & 59.3          & 66.1          & 79.8          & 84.9          & 88.3          \\
\multicolumn{1}{c|}{} 
                             & Ours                        & \textbf{23.9} & \textbf{44.7} & \textbf{58.8} & 64.7          & 69.4          & \textbf{60.7} & \textbf{68.8} & \textbf{82.2} & \textbf{86.5} & \textbf{90.3} \\ \hline
\multicolumn{1}{c|}{}        & EUG \cite{wu2018exploit}    & 26.6          & 48.7          & 63.4          & -             & 72.6          & 59.5          & 69.0          & 81.0          & -             & 89.5          \\
\multicolumn{1}{c|}{ER=20\%} & PL\cite{wu2019progressive}  & 27.2          & 49.6          & \textbf{64.5} & 69.8          & 74.4          & 59.6          & 69.1          & 81.2          & 85.6          & 89.6          \\
\multicolumn{1}{c|}{}        & Ours                        & \textbf{32.1} & \textbf{51.1} & 64.4          & \textbf{70.4} & \textbf{75.6} & \textbf{62.2} & \textbf{70.2} & \textbf{84.8} & \textbf{89.0} & \textbf{92.3} \\ \hline
\multicolumn{1}{c|}{}        & EUG\cite{wu2018exploit}     & 29.6          & 52.3          & 64.2          & -             & 73.1          & 59.2          & 69.1          & 81.2          & -             & 88.9          \\
\multicolumn{1}{c|}{ER=15\%} & PL\cite{wu2019progressive}  & 29.9          & 52.7          & 66.3          & 71.9          & 76.4          & 59.5          & 69.3          & 81.4          & 85.9          & 89.2          \\
\multicolumn{1}{c|}{}        & Ours                        & \textbf{34.3} & \textbf{54.5} & \textbf{66.8} & \textbf{72.2} & \textbf{77.2} & \textbf{63.8} & \textbf{71.9} & \textbf{84.8} & \textbf{87.3} & \textbf{91.0} \\ \hline
\multicolumn{1}{c|}{}        & EUG\cite{wu2018exploit}     & 34.7          & 57.6          & 69.6          & -             & 78.1          & 61.8          & 70.8          & 83.6          & -             & 89.6          \\
\multicolumn{1}{c|}{ER=10\%} & PL\cite{wu2019progressive}  & 34.9          & 57.9          & 70.3          & 75.2          & 79.3          & 61.9          & 71.0          & 83.8          & 87.4          & 90.3          \\
\multicolumn{1}{c|}{}        & Ours                        & \textbf{43.0} & \textbf{60.9} & \textbf{75.8} & \textbf{80.3} & \textbf{84.8} & \textbf{64.4} & \textbf{71.1} & \textbf{86.5} & \textbf{89.3} & \textbf{92.5} \\ \hline
\multicolumn{1}{c|}{}        & EUG\cite{wu2018exploit}     & 42.5          & 62.7          & 74.9          & -             & 82.6          & 63.2          & 72.8          & 84.2          & -             & 91.5          \\
\multicolumn{1}{c|}{ER=5\%}  & PL\cite {wu2019progressive} & 42.6          & 62.8          & 75.2          & 80.4          & 83.8          & 63.3          & 72.9          & 84.3          & 88.3          & 91.4          \\
\multicolumn{1}{c|}{}        & Ours                        & \textbf{46.3} & \textbf{66.7} & \textbf{78.1} & \textbf{81.6} & \textbf{85.8} & \textbf{65.0} & \textbf{73.3} & \textbf{84.8} & \textbf{89.0} & \textbf{91.6} \\ \Xhline{1.3pt}
\end{tabular}

\
\setlength{\abovecaptionskip}{1pt} 

\label{tab Table 2}
\vspace{-0.2cm}
\end{table*}

We analyze the characteristics of samples in the DukeMTMC-reID dataset to find the reasons for this phenomenon. We found there are considerable differences among the in-class samples. Researchers collect samples in the winter from $8$ camera views. The same identity pedestrian wears a coat and hat under the current camera and takes off his coat when he came to another camera. Samples like this are prevalent in the DukeMTMC-reID dataset. The method of uncertainty estimation to train the network in one-shot scenarios with small differences between classes and large inter-class differences has significant limitations, which leads to a terrible beginning. The ideal experimental effect cannot be achieved in the later iterative training. According to the analysis above, we suggest that the network set a larger Expansion Rate to avoid the trap of local uncertainty estimation in the case of large differences between sample classes. In the case of high-similarity samples, the network needs to set a smaller Expansion Rate in order to obtain better performance.

The performance of our proposed method is shown in Table~\ref{tab Table 1} and Table~\ref{tab Table 2}. Experimental results show that our method achieves satisfactory performance, using only one labeled sample of each identity on four large-scale datasets.
%-------------------------------------------------------------------------

\subsection{Ablation study}
\label{sec:abl_study}

\begin{figure}[t] 
\begin{center}
%\fbox{\rule{0pt}{2in} \rule{1\linewidth}{0pt}}
\includegraphics[width=0.8\linewidth]{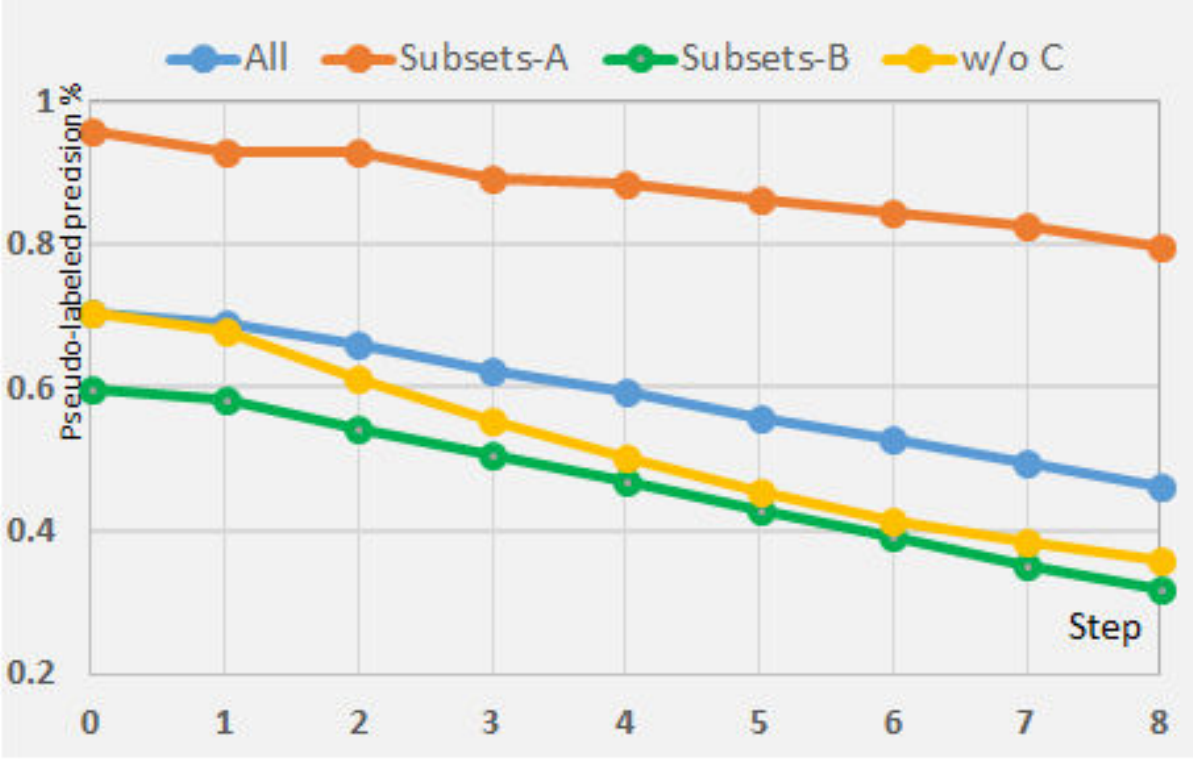}
\end{center}
\setlength{\abovecaptionskip}{1pt} 
   \caption{The precision of selected pseudo-labeled samples during iterations.}
\label{fig:10}

\end{figure}

In this subsection, in order to show the effectiveness of the SPUE-Net, we give an ablation study to evaluate different variants of the proposed method for Market-1501 of image-based and MARS of video-based datasets are employed. We conduct ablation studies on two crucial parts of SPUE-Net, ie, uncertainty estimation and the Co-operative learning method. as shown in Table~\ref{tab Table 3} and Figure~\ref{fig:4}. We set the same initial labeled image and training settings in all experiments. “Ours w/o C" indicates that we only use the uncertainty estimation without the Co-operative learning method. “Ours w/o C\&U" is consistent with the experimental settings of the baseline\cite{ wu2019progressive} without uncertainty estimation and the Co-operative learning method.

\begin{table}[]
  \centering
  \Large

\setlength{\tabcolsep}{0.5mm}
\caption{Ablation study compares two criteria on Market-1501 of image-based and MARS of video-based datasets.}

\begin{tabular}{cl!{\vrule width1.1pt}cc|cc}
\Xhline{1.1pt}
\multirow{2}{*}{setting}                    & \multicolumn{1}{c!{\vrule width1.1pt}}{\multirow{2}{*}{Method}} & \multicolumn{2}{c|}{Market-1501} & \multicolumn{2}{c}{MARS} \\ \cline{3-6} 
                                            & \multicolumn{1}{c!{\vrule width1.1pt}}{}                        & mAP            & Rank-1          & mAP        & Rank-1       \\  \Xhline{1.1pt}
\multirow{3}{*}{ER=30\%}                    & Ours w/o C\&U                                & 13.4           & 35.5            & 22.1       & 44.5         \\
                                            & Ours w/o C                                   & 10.8           & 32.9            & 20.0       & 39.5         \\
                                            & Ours                                         & 16.2           & 38.5            & 23.9       & 44.7         \\ \hline
\multirow{3}{*}{ER=20\%}                    & Ours w/o C\&U                                & 17.4           & 41.4            & 27.2       & 49.6         \\
                                            & Ours w/o C                                   & 15.1           & 38.2            & 24.9       & 45.6         \\
                                            & Ours                                         & 19.6           & 45.0            & 32.1       & 51.1         \\ \hline
\multirow{3}{*}{ER=10\%}                    & Ours w/o C\&U                                & 23.2           & 51.5            & 34.9       & 57.9         \\
                                            & Ours w/o C                                   & 18.6           & 45.2            & 33.4       & 52.1         \\
                                            & Ours                                         & 28.3           & 57.5            & 43.0       & 60.9         \\ \hline
\multicolumn{1}{l}{\multirow{3}{*}{ER=5\%}} & Ours w/o C\&U                                & 26.2           & 55.8            & 42.6       & 62.8         \\
\multicolumn{1}{l}{}                        & Ours w/o C                                   & 20.2           & 51.4            & 38.5       & 58.8         \\
\multicolumn{1}{l}{}                        & Ours                                         & 30.5           & 60.4            & 46.3       & 66.7         \\ \Xhline{1.1pt}
\end{tabular}

\
\setlength{\abovecaptionskip}{10pt} 

\label{tab Table 3}
\vspace{-0.2cm}
\end{table}

\subsubsection{The limitations of uncertainty estimation} 
We compare our method to the model trained without the Co-operative learning method, denoted as “Ours w/o C” in Table~\ref{tab Table 3}. The network only uses the uncertainty estimation to have a certain level of decline relative to the baseline \cite{wu2019progressive} under different ER settings. However, we can observe that the performance of the model "Ours w/o C" is the best in the early iterations(from 1 to 3), proving the validity of the uncertainty estimation in Figure~\ref{fig:4}. Since the network only selects a few of the most reliable unlabeled samples as pseudo-labeled samples, the accuracy score of label estimation is relatively high at the beginning. As the number of iterations increases, the number of pseudo-label samples selected by the network increases, while the accuracy of pseudo-label samples continues to decrease, as shown in the blue curve in Figure~\ref{fig:10}. The blue curve is the change of the precision of all pseudo-label samples in each step when the ER is set to 10 in the market-1501 dataset. After the middle of the iteration, the Re-id evaluation performance dropped significantly from the baseline\cite{wu2019progressive}. From the perspective of experimental research, uncertainty estimation improves robustness by assigning large variances to mislabeled samples, reducing the influence of mislabeled samples on model training, which obtain a better embedding space and different identities become more separable. However, when the number of mislabeled samples involved in uncertainty estimation is relatively large, common features of inner-class samples are corroded. The uncertainty estimated by the model will have a negative impact on the network.

\subsubsection{The effectiveness of the Co-operative learning method} 
\begin{figure}[t]  
\begin{center}
%\fbox{\rule{0pt}{2in} \rule{1\linewidth}{0pt}}
\includegraphics[width=1\linewidth]{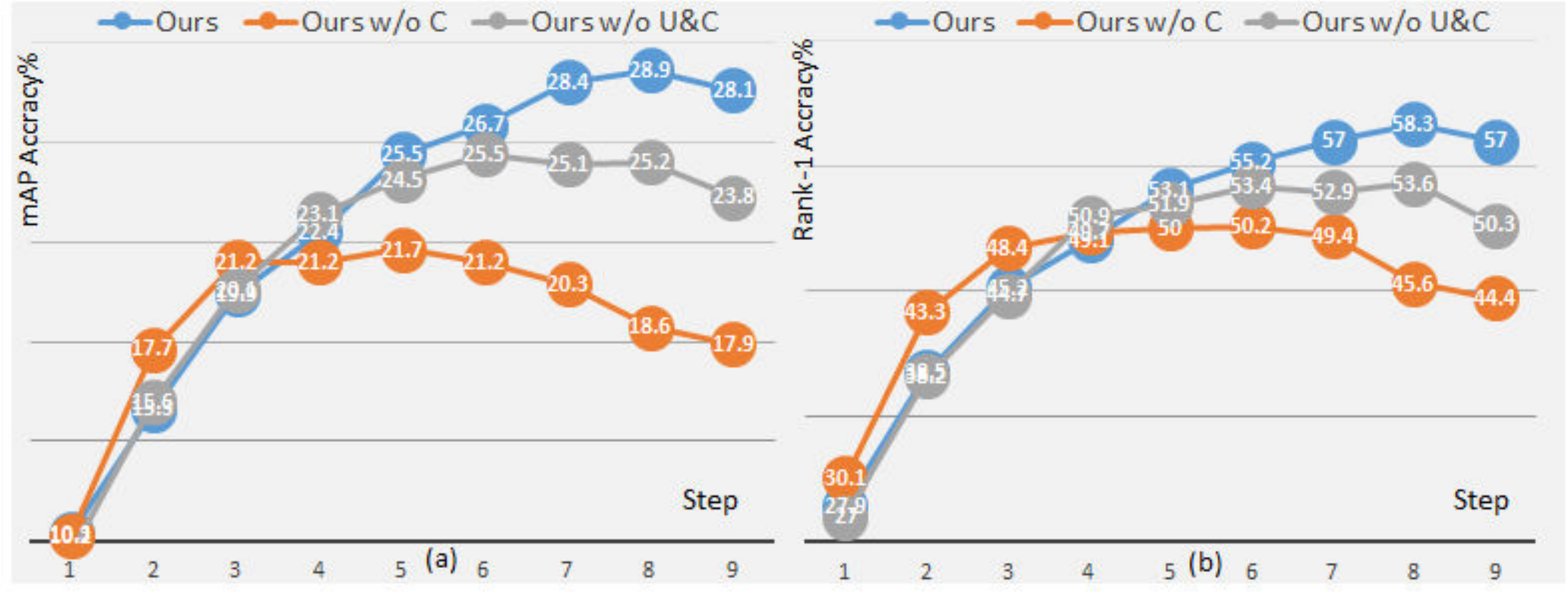}
\end{center}
\setlength{\abovecaptionskip}{1pt} 
   \caption{Ablation studies on Market-1501. (a) and (b) are mAP and Rank-1 on the evaluation set during iterations. The expansion rate is ER=10\%. The x-axis stands for the iterative steps.}
\label{fig:4}
\vspace{-0.2cm}
\end{figure}

The Re-ID performances of sampling by uncertainty estimation and the Co-operative learning method are illustrated in Figure~\ref{fig:4} and Table~\ref{tab Table 3}. For training that only uses uncertainty estimation, there are a large number of mislabeled samples in the pseudo-labeled samples predicted by the uncertainty estimation at the later stage of the iteration from Figure~\ref{fig:10}, which affects the uncertainty model's correct estimation of sample noise, resulting in a decrease in model performance. The SPUE-Net adopts the Co-operative learning method of local uncertainty estimation and determinacy estimation. Through the subsets division based on the distance confidence, the uncertainty estimation and the determinacy estimation have a positive incentive effect, resulting in the model that can alleviate the corrosion of inherent features, achieve further feature mining and improve the performance of the model.

%-------------------------------------------------------------------------
\subsection{Algorithm analysis} 
\label{sec:fes}

\begin{figure}[t] 
\begin{center}
%\fbox{\rule{0pt}{2in} \rule{1\linewidth}{0pt}}
\includegraphics[width=1\linewidth]{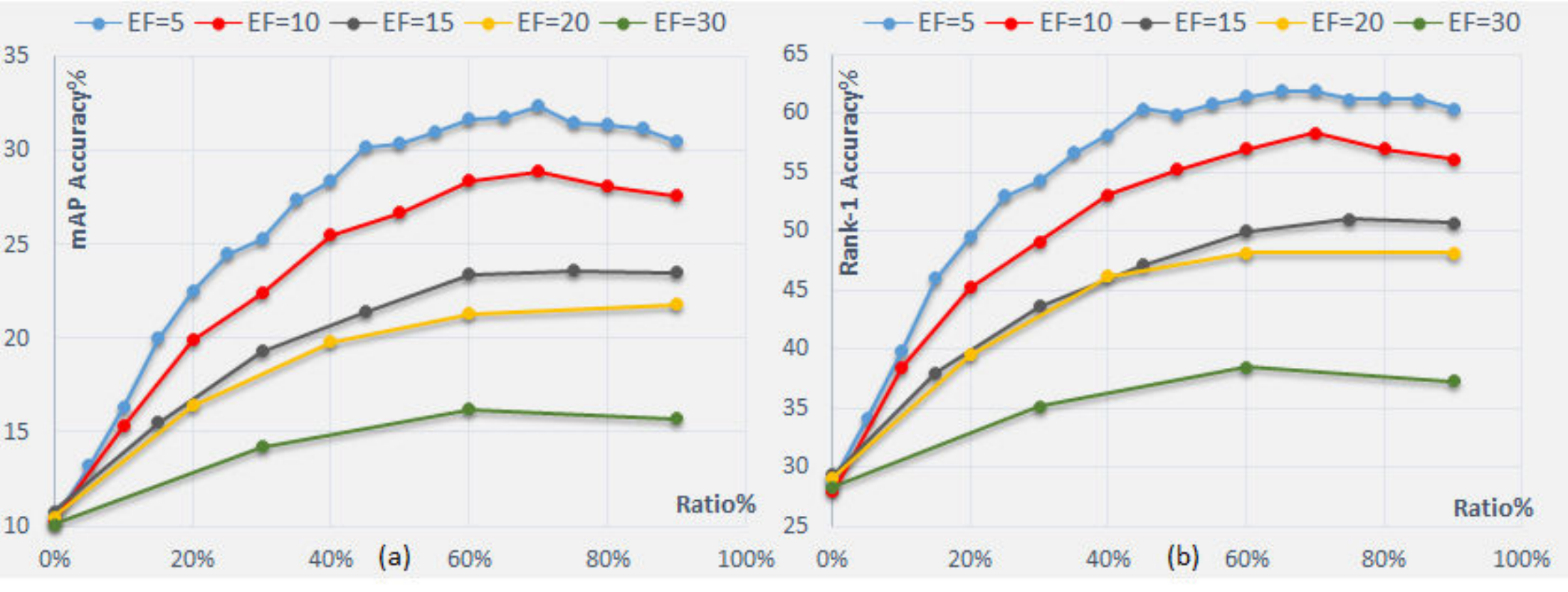}
\end{center}
\setlength{\abovecaptionskip}{1pt} 
   \caption{Comparison with different values of expansion rate on Market-150. “ER” denotes the expansion rate. Fig (a) and fig (b) are mAP and Rank-1 accuracy on the evaluation set of the different expansion rates during iterations. Each solid point indicates an iteration step. The x-axis stands for the ratio of the selected sample from unlabeled data.}
\label{fig:5}
\vspace{-0.2cm}
\end{figure}
%-------------------------------------------------------------------------
 
\subsubsection{Expansion Rate and Iterations Analysis} There are two crucial parameters in the self-paced sampling strategy, ie, Expansion Rate (ER) and Iterations, which determine the training speed and performance of the model. ER represents the expansion speed of the pseudo-labeled dataset during iterations. The change of the ER affects the number of iterations, and which is in an inverse proportion. The results of different ER on Market-1501 are shown in Figure~\ref{fig:5}. Figure ~\ref{fig:5}(a) and Figure~\ref{fig:5}(b) are mAP and Rank-1 scores on the evaluation set with the different expansion rates, respectively.  The experimental scores of EF=5\% are the best. The experimental results show that a smaller ER yields a better performance of the model. We observe from Figure~\ref{fig:5} that as the number of iterations increases, the gaps among the five curves continue to increase, which indicates that the error in pseudo-labeled estimation increases.  

Figure~\ref{fig:5} shows the experimental results of each iteration. Throughout iterations, the precision of the mAP and Rank-1 scores on Market-1501 continues to increase, which validates the stability of our model. However, the evaluation performance stops to increase in the last few iterations. The gain of adding new pseudo-labeled samples is consumed by the larger number of mislabeled samples. The mislabeled samples in the inchoate iterations of the self-paced sampling will have the most severe impact on the later iterative process. The uncertainty estimated by the model will harm the network. If the space covered by labeled samples is inconsistent with the real data space, the self-paced training classification algorithm will be completely ineffective, and the unlabeled samples are not conducive to learn a better classifier.

%-------------------------------------------------------------------------

%-------------------------------------------------------------------------

\begin{figure}[t] 
\begin{center}
%\fbox{\rule{0pt}{2in} \rule{1\linewidth}{0pt}}
\includegraphics[width=1\linewidth]{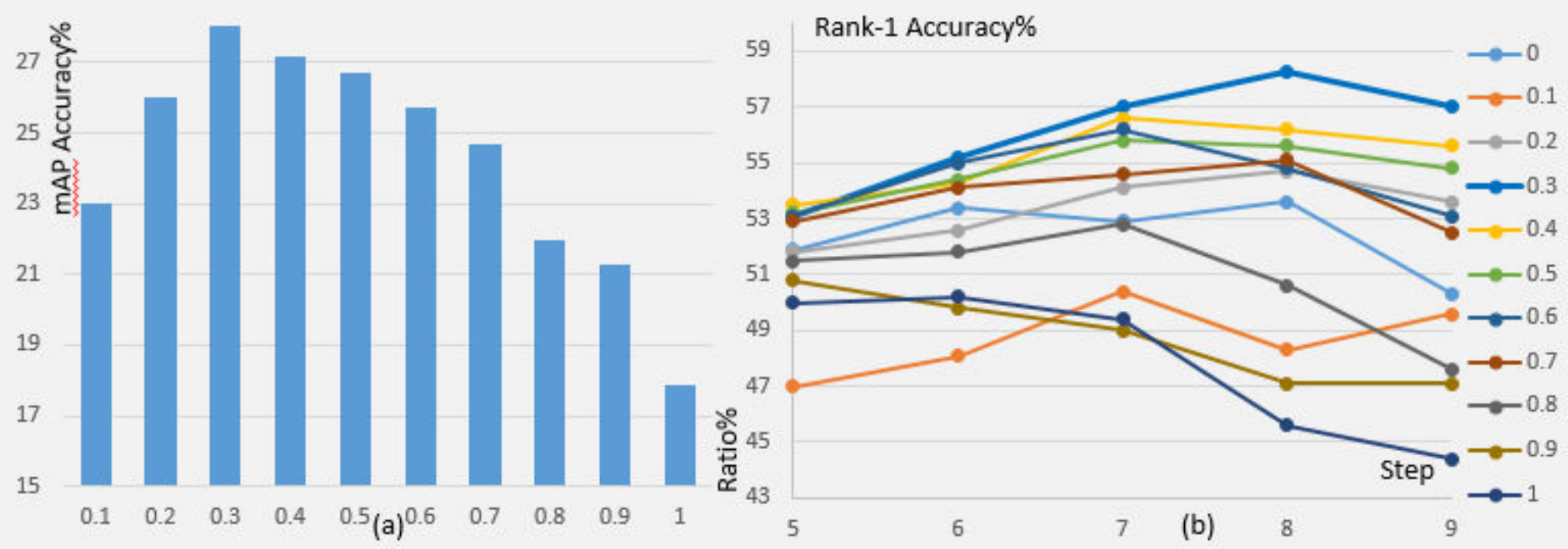}
\end{center}
\setlength{\abovecaptionskip}{1pt} 
   \caption{Comparison with different values of hyper-parameter $\alpha$ on Market-150. Fig (a) is mAP accuracy on the evaluation set of the different hyper-parameter $\alpha$ in the last iteration, and Fig (b) is Rank-1 accuracy on the evaluation set of the different hyper-parameter $\alpha$ during iterations.}
\label{fig:8}
\vspace{-0.2cm}
\end{figure}

\subsubsection{Subsets division Analysis} We compare the different values of hyperparameter $\alpha$, and it adjusts the proportion of Subset-$\mathcal{A}$ and Subset-$\mathcal{B}$ in the pseudo-label dataset. Figure ~\ref{fig:8} shows the effect of different hyperparameters $\alpha$ on the performance of RE-ID model. To visualize the experimental results more clearly, we show the final step of Rank-1 (iteration 5 to 9) in Figure~\ref{fig:8} (b). As we can see, $\alpha=0.3$ performs best in the experimental evaluation of the mAP and Rank-1. In this way, Subset-$\mathcal{A}$ selects samples with confidence in the first 30\% to learn the feature and uncertainty, and Subset-$\mathcal{B}$ chooses the remaining 70\% samples to learn deterministic feature embedding. The uncertainty estimation method pursues a balanced relationship between the number of training samples and the performance of the re-id model. Co-operative learning of labeled samples, pseudo-labeled samples(Subset-$\mathcal{A}$ and Subset-$\mathcal{B}$), and unlabeled samples make the experimental result the best performance.
%-------------------------------------------------------------------------

\section{Uncertainty Thinking}
The difference between the uncertainty estimation and the determinacy estimation is that the sample feature and noise are estimated simultaneously. The estimated uncertainty is closely related to the quality of pedestrian images, that is, the uncertainty of learning increases as the image quality decreases. The uncertainty estimation of samples can improve the learning of identity features by adaptively reducing the adverse effects of noisy training samples, which makes the high-quality (low-noise) samples more compact and low-quality (high-noise) samples farther away. Uncertainty estimation reduces the interference of low-quality samples to the model and increases the training weight of high-quality samples. Accurately estimated noise is the important core of uncertainty estimation.

What is sample noise? The model has no labels for noise and features and estimates that the sample noise comes from a random sampling of hidden space features. Under the supervision of the identity labels, the model gradually learns the common features $\mu$ and difference features $\sigma$ of the same class samples. Therefore, when the mislabeled samples account for most of the training samples, the features and noise learned by the model will have a greater deviation. It is even more serious than the deviation of the deterministic representation, which has been verified in the ablation study\ref{sec:abl_study}. When high-quality samples dominate the training samples, the acquired features and noise are close to real values, thereby reducing the adverse effects of mislabeled samples on the model. When the number of samples is small and the differences between samples are large, the features and noise learned by the model tend to be unstable. This is also an important reason for the decline in the first iteration of the DukeMTMC-reID training process. [see details in Section\ref{sec:cmp_state}]. In general, it is not appropriate to use uncertainty estimation in samples with poor data quality. Uncertainty estimation can improve the effect of model performance with relatively better quality samples.

%-------------------------------------------------------------------------

\section{Conclusion}
In this paper, we propose a Self-Paced Uncertainty Estimation Network (SPUE-Net) for one-shot person Re-ID. With the self-paced learning, subsets division strategy, and the Co-operative learning method of local uncertainty estimation and deterministic estimation, our method is able to capture and remove the model uncertainty and data uncertainty without extra supervision information. The performance improvement demonstrates that our proposed SPUE-Net performs better than deterministic models on most benchmarks.  
%-------------------------------------------------------------------------

\normalem
\bibliographystyle{ACM-Reference-Format}
\bibliography{SPUE}

\end{document}